\documentclass[runningheads]{llncs}
\usepackage{xcolor}
\usepackage[T1]{fontenc}
\usepackage{graphicx}
\usepackage{amssymb}
\usepackage{marvosym}
\usepackage{bm}
\usepackage{amsmath}
\usepackage{diagbox}
\usepackage{subfigure}
\usepackage{xcolor}
\usepackage{multirow}
\usepackage{subcaption}
\usepackage{booktabs}
\usepackage{hyperref}
\usepackage{url}
\usepackage{verbatim}

\usepackage{multirow}

\begin{document}
\title{Adapting Foundation Model for Dental Caries Detection with Dual-View Co-Training}
\titlerunning{DVCTNet: Dual-View Co-Training Network}

\author{
Tao Luo \inst{1}\textsuperscript{*} \and
Han Wu\inst{1,2}\textsuperscript{*} \and
Tong Yang\inst{5} \and 
Dinggang Shen\inst{1,3,4} \and
Zhiming Cui\inst{1}\textsuperscript{(\Letter)}}

\authorrunning{T Luo et al.}
\institute{School of Biomedical Engineering \& State Key Laboratory of Advanced Medical Materials and Devices, ShanghaiTech University, Shanghai, China\\
\email{cuizhm@shanghaitech.edu.cn}\and
Lingang Laboratory, Shanghai, China\and
Shanghai United Imaging Intelligence Co. Ltd., Shanghai, China\and
Shanghai Clinical Research and Trial Center, Shanghai, China \and
Shanghai Linkedcare Information Technology Co., Ltd., Shanghai, China}

\footnotetext{* indicates equal contribution.}
\maketitle
\begin{abstract}
Accurate dental caries detection from panoramic X-rays plays a pivotal role in preventing lesion progression.
However, current detection methods often yield suboptimal accuracy due to subtle contrast variations and diverse lesion morphology of dental caries.
In this work, inspired by the clinical workflow where dentists systematically combine whole-image screening with detailed tooth-level inspection, we present \textbf{DVCTNet}, a novel \underline{\textbf{D}}ual-\underline{\textbf{V}}iew \underline{\textbf{C}}o-\underline{\textbf{T}}raining network for accurate dental caries detection.
Our DVCTNet starts with employing automated tooth detection to establish two complementary views: a global view from panoramic X-ray images and a local view from cropped tooth images. We then pretrain two vision foundation models separately on the two views.
The global-view foundation model serves as the detection backbone, generating region proposals and global features, while the local-view model extracts detailed features from corresponding cropped tooth patches matched by the region proposals.
To effectively integrate information from both views, we introduce a Gated Cross-View Attention (GCV-Atten) module that dynamically fuses dual-view features, enhancing the detection pipeline by integrating the fused features back into the detection model for final caries detection.
To rigorously evaluate our DVCTNet, we test it on a public dataset and further validate its performance on a newly curated, high-precision dental caries detection dataset, annotated using both intra-oral images and panoramic X-rays for double verification.
Experimental results demonstrate DVCTNet's superior performance against existing state-of-the-art (SOTA) methods on both datasets, indicating the clinical applicability of our method.
Our code and labeled dataset are available at \href{https://github.com/ShanghaiTech-IMPACT/DVCTNet}{https://github.com/ShanghaiTech-IMPACT/DVCTNet}.


\end{abstract}

\section{Introduction}
Dental caries is one of the most common oral diseases~\cite{jain2024s,wen2022global}, which may lead to irreversible teeth damage. Accurate caries detection from panoramic X-rays is crucial to prevent lesion progression~\cite{mei2023hc,haghanifar2023paxnet}. 
However, manually detecting caries from panoramic X-rays remains time-consuming and labor-intensive due to their radiographic diversity and inconspicuous signs.
With the development of deep learning~\cite{cui2022fully,wu2024cephalometric,zhao2025clip}, existing methods have shown great success in automated caries detection from panoramic X-rays~\cite{imak2022dental,zhang2022development,mohammad2022deep}. For example, Zhu et al.~\cite{zhu2023cariesnet} incorporated a full-scale axial attention module, achieving notable performance in segmenting caries.  
Wang et al.~\cite{wang2023deep} proposed a lightweight region of interest pruning method that effectively improves caries detection with a novel label assignment head.
Chen et al.~\cite{chen2024cariesxrays}(referred to as FPCL) proposed a proposal-prototype contrastive learning method based on a bidirectional FPN for caries detection. 

\begin{figure}[t!]
    \centering
    \includegraphics[width=\textwidth, height=75pt]{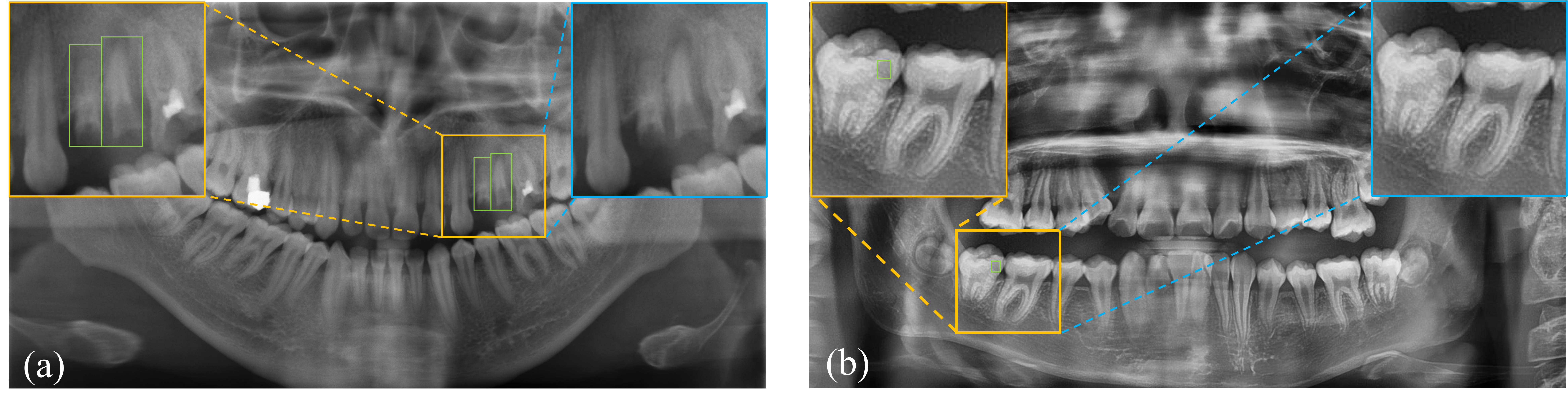}
    \caption{Comparison of annotated cases between recent public dataset ~\cite{chen2024cariesxrays} (a), and our dataset (b). Dashed lines link zoomed views: blue section shows the original image, orange section highlights the same area with overlaid annotated boxes.}
    \label{fig:compare}
\end{figure}

Despite encouraging performance, current approaches still face two major limitations.
1) Learning-based methods rely heavily on the existence of well-annotated datasets, while current public datasets~\cite{chen2024cariesxrays,zhang2023children} fail to provide a high-quality golden-standard dataset where clinical practice addresses uncertain dental caries through dual-modality verification with panoramic X-rays and intra-oral images~\cite{park2022caries}. Additionally, they tend to over-annotate caries areas as shown in Fig. \ref{fig:compare}.
2) Existing methods directly adapt generic models from computer vision, which are either limited to single-view analysis at the image level or fail to align with the comprehensive clinical diagnostic workflow, leading to suboptimal results~\cite{karakucs2024ai,wang2023deep}.

In this paper, we introduce DVCTNet, a novel dual-view co-training framework that leverages foundation models for dental caries detection. Specifically, DVCTNet generates dual-view images: panoramic X-rays and cropped tooth images, to align with the clinical workflow, where dentists typically combine whole-image screening with detailed tooth-level inspection for accurate diagnosis.
Our DVCTNet pretrains two foundation models on global and local views, yielding dual-view image encoders on the collected large-scale, unlabeled dataset. The global view encoder serves as the backbone for the object detector, generating region proposals and global features, while the local view encoder extracts fine-grained features from cropped tooth patches matched with the region proposals, complementing the global representation. 
To effectively integrate dual-view features, a Gated Cross-View Attention (GCN-Atten) mechanism is developed to dynamically fuse the features, enhancing the detection pipeline by reincorporating the fused features into the model for final caries detection.
We conducted extensive experiments on two datasets: 1) a publicly available dataset from~\cite{chen2024cariesxrays}, and 2) a high-quality and newly-collected dataset dual-verified with intra-oral images and panoramic X-rays, and cross-validated by four dental experts. Experimental results demonstrate that DVCTNet outperforms existing state-of-the-art (SOTA) methods on both datasets, indicating strong clinical applicability.
In summary, \textbf{our contributions can be summerized as followed}: 
1) A novel dual-view co-training mechanism that leverages foundation models to extract and utilize complementary features from global panoramic and local tooth-specific views for accurate dental caries detection.
2) A Gated Cross-View Attention mechanism that dynamically fuses features from dual views, optimizing the integration of contextual and detail-oriented information.
3) We present the first high-precision benchmark dataset for dental caries detection with annotations double-verified through intra-oral images and panoramic X-rays.

\section{Method}
\begin{figure}[t!]
    \centering
    \includegraphics[width=\textwidth]{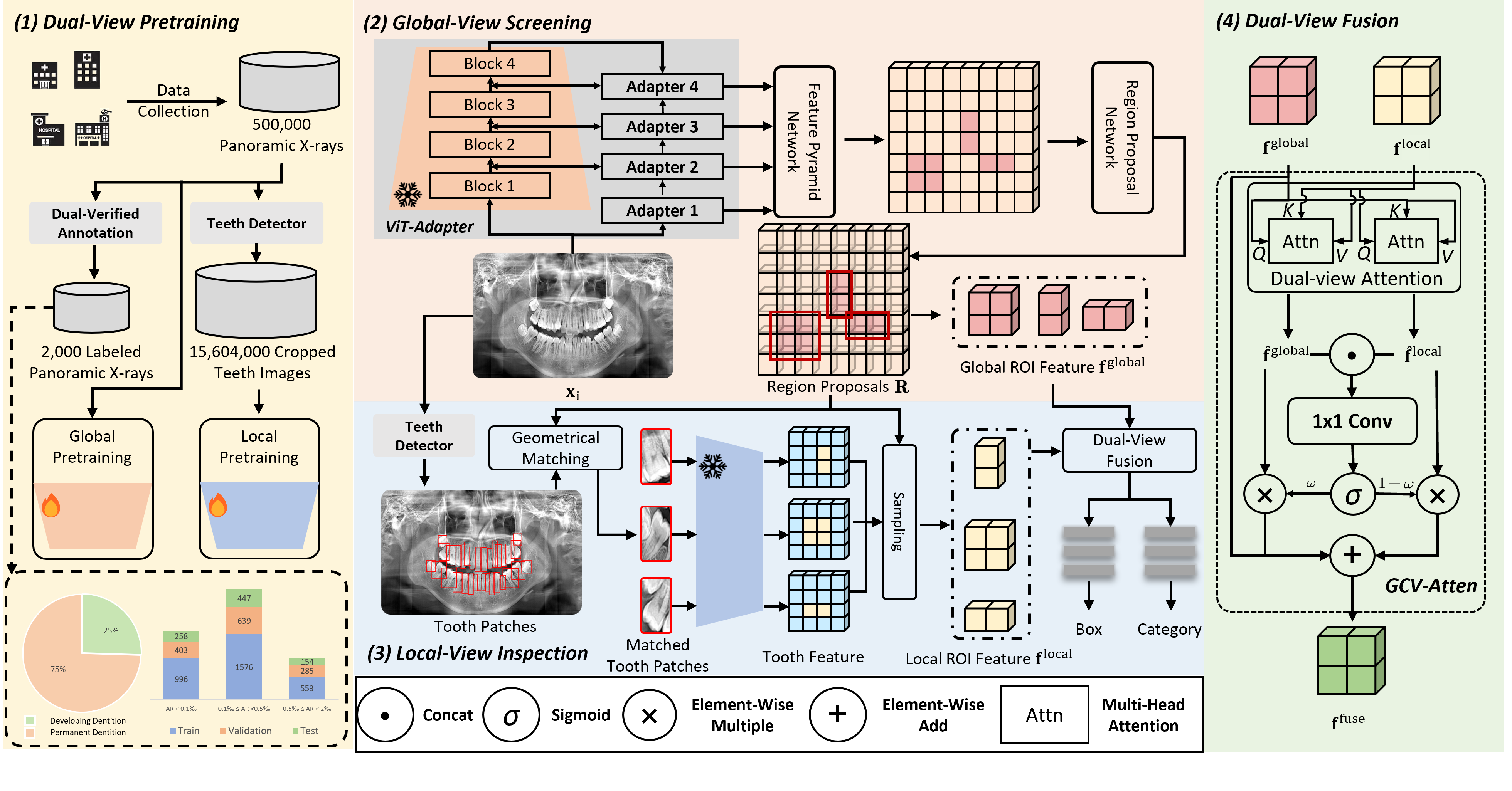}
    \caption{An overview of the proposed DVCTNet for dental caries detection.}
    \label{fig:pipeline}
\end{figure}

The overview pipeline of DVCTNet is shown in Fig. \ref{fig:pipeline}. First, we establish complementary views and pretrain foundation models on both panoramic X-rays and tooth images (Sec. \ref{Dual-viewPretraining}). Next, we integrate these pretrained models through dual-view co-training for caries detection, combining global screening with local inspection (Sec. \ref{Dual-coTraining}). Finally, we introduce a Gated Cross-View Attention mechanism for dual-view fusion (Sec. \ref{Dual-viewFusion}).

\subsection{Dual-View Pretraining}\label{Dual-viewPretraining}

Our method begins with the establishment of dual views that closely follow the clinical diagnostic workflow for caries detection. Given an input panoramic X-ray image $\mathbf{X} \in \mathbb{R}^{H \times W}$, we employ a well-developed tooth detector $\mathcal{D}$ (i.e., tooth detector form ~\cite{mei2023hc}) to generate bounding boxes around individual teeth and crop each tooth, resulting in local tooth images $\mathbf{T}=\{ \mathbf{t}_i \in \mathbb{R}^{h \times w} \}_{i=1}^{N}$, where $N$ is the number of teeth present in the X-ray. $H \times W$ and $h \times w$ are the resolutions of the X-ray and cropped tooth image, respectively.
This process yields dual views with complementary information: a global view of the entire panoramic X-ray image $\mathbf{X}$ and a corresponding local view $\mathbf{T}$ consisting of individual tooth images.

As shown in Fig.\ref{fig:pipeline}(1), to effectively utilize the large amount of unlabeled data available in dental imaging and to better capture the intrinsic features of both global and local views, we introduce a dual-view pretraining stage based on DINOv2~\cite{oquab2024dinov2}, a state-of-the-art self-supervised learning framework. DINOv2 employs a teacher-student distillation mechanism with momentum updating, allowing the model to learn consistent representations across different augmentations of the same image. 
In our study, for the \textbf{global view pretraining}, we configure the ViT-B model with a patch size of $14\times14$ and input resolution of $518\times 518$ pixels. During training, the model uses multi-scale argumentation with different crops, e.g., $518\times 518$ and $224\times 224$ pixels, enabling the model to capture hierarchical information in the panoramic X-rays. For the \textbf{local view pretraining}, we use the ViT-B model but with smaller crop dimensions appropriate for tooth-level analysis, e.g., $112\times 112$ and $98\times 98$ pixels. This configuration is specifically designed to effectively capture the fine-grained dental information presented in individual tooth images.

\subsection{Dual-View Co-Training}\label{Dual-coTraining}
Following the dual-view pretraining stage, we integrate the pretrained models into dental caries detection. The core idea of the dual-view co-training is to simulate the real-world clinical workflow, where dentists combine global-view screening from the entire panoramic X-ray images with local-view inspection at the tooth level for an accurate diagnosis.

\subsubsection{Global-View Screening} As shown in Fig. \ref{fig:pipeline}(2), we develop a multi-scale feature extraction framework based on ViTAdapter~\cite{chen2022vision}. The ViTAdapter extends the original Vision Transformer architecture by introducing deformable attention modules between transformer blocks, enabling the model to capture multi-scale features more effectively. Specifically, we apply these adapters among the original ViT blocks of the image-level encoder pretrained in Sec.\ref{Dual-viewPretraining}. This design allows our model to generate multi-scale feature maps with dimensions consistent with traditional detection backbones~\cite{he2016deep}. These feature maps containing hierarchical semantic information are then fused through a Feature Pyramid Network (FPN) to combine multi-scale information. A Region Proposal Network (RPN) then generates region proposals $\mathbf{R}=\{\mathbf{r}_j, \mathbf{f}_j^{\mathrm{global}} \}_{j=1}^{M}$ from these fused features, where each $\mathbf{r}_j \in \mathbb{R}^4$ represents the bounding box of the proposal (center coordinates, width, and height), and $\mathbf{f}_j^{\mathrm{global}} \in \mathbb{R}^{7\times7}$ indicates the corresponding ROI features form the global view using RoIAlign. 

\subsubsection{Local-View Inspection}As shown in Fig. \ref{fig:pipeline}(3), to ensure our model simultaneously capture the fine-grained details from individual teeth, we perform geometrical matching between each region proposal bounding box $\mathbf{r}_{j}$ and the most appropriate tooth image $\mathbf{t}_{i^*} \in \mathbf{T}$ detected from the entire panoramic X-ray. The matching is based on Intersection over Detection (IoD) overlap, defined as:
\begin{equation}
    i^* = \arg\max_{\mathbf{t}_i \in \mathbf{T}} \text{IoD}(\mathbf{r}_j, \mathcal{B}(\mathbf{t}_i)),
\end{equation}
where $\mathcal{B}(\mathbf{t}_i)$ denotes the bounding box of the cropped tooth image $\mathbf{t}_i$ in the original panoramic X-ray coordinate system, and IoD is calculated as:
\begin{equation}
    \text{IoD}(\mathbf{r}_j, \mathcal{B}(\mathbf{t}_i)) = \frac{\text{Area}(\mathbf{r}_j \cap \mathcal{B}(\mathbf{t}_i))}{\mathcal{B}(\mathbf{t}_i)}.
\end{equation}
Once the matching tooth image $\mathbf{t}_{i^*}$ is identified, it is passed through the pretrained local-view encoder (introduced in Sec.\ref{Dual-viewPretraining}) to extract detailed tooth-level features, and crop the corresponding ROI feature $\mathbf{f}_{i^*}^{\mathrm{local}} \in \mathbb{R}^{7\times7}$ from the local view.

\subsection{Dual-View Fusion}\label{Dual-viewFusion}
To effectively integrate the complementary information from both views, we use a Gated Cross-View Attention (GCV-Atten) fusion module. This module dynamically fuses features from the global and local views based on their relevance to caries detection.

As illustrated in Fig. \ref{fig:pipeline}(4), given a ROI feature $\mathbf{f}_j^{\mathrm{global}}$ from the global view and $\mathbf{f}_{i^*}^{\mathrm{local}}$ from the local view, we first compute a dual-direction scaled dot-product attention mechanism between them. This allows each view to attend to relevant information in the other view. The attention mechanism is formulated as:
\begin{align}
    \mathbf{\hat{f}}_j^{\mathrm{global}} &= \mathrm{Softmax}\left(\frac{\mathbf{f}_j^{\mathrm{global}} \cdot {\mathbf{f}_{i^*}^{\mathrm{local}}}^\mathrm{T}}{\sqrt{d_k}}\right) \cdot \mathbf{f}_j^{\mathrm{global}}, \\
    \mathbf{\hat{f}}_{i^*}^{\mathrm{local}} &= \mathrm{Softmax}\left(\frac{\mathbf{f}_{i^*}^{\mathrm{local}} \cdot {\mathbf{f}_j^{\mathrm{global}}}^\mathrm{T}}{\sqrt{d_k}}\right) \cdot \mathbf{f}_{i^*}^{\mathrm{local}},
\end{align}
where $\mathbf{\hat{f}}_j^{\mathrm{global}}$ and $\mathbf{\hat{f}}_{i^*}^{\mathrm{local}}$ are the attention-weighted features, and $\sqrt{d_k}$ is the scaling factor. After computing the attention-weighted features, we concatenate them and pass them through a 1D convolutional layer to reduce the channel dimension. Then, a sigmoid activation is followed to compute an attention weight $\omega$:
\begin{equation}
    \omega = \mathrm{Sigmoid}(\mathrm{Conv1D}(\mathrm{Concat}(\mathbf{\hat{f}}_j^{\mathrm{global}}, \mathbf{\hat{f}}_{i^*}^{\mathrm{local}}))).
\end{equation}
In this way, the final fused ROI features can be computed as a weighted combination of the attention-weighted features from both views, with an additional residual connection from the global view features $\mathbf{f}_j^{\mathrm{global}}$:
\begin{equation}
    \mathbf{f}_j^{\mathrm{fuse}} = \omega \cdot \mathbf{\hat{f}}_j^{\mathrm{global}} + (1-\omega) \cdot \mathbf{\hat{f}}_{i^*}^{\mathrm{local}} + \mathbf{f}_j^{\mathrm{global}}.
\end{equation}
The residual connection from the global view ensures the global contextual information is preserved even after fusion. And the weighted combination allows the model to adaptively balance the contributions from both views based on their relevance to the current region. Finally, the fused feature $\mathbf{f}_j^{\mathrm{fuse}}$ is fed back to the detection head for final dental caries classification and bounding box regression.

\section{Experiments}
\subsection{Dataset and Evaluation Metrics}
\subsubsection{AAAI Dataset}
FPCL~\cite{chen2024cariesxrays} collected and released a CariesXrays dataset which covers 6,000 panoramic dental X-ray images, with a total of 13,783 instances of dental caries. The whole dataset is randomly divided into 4,800/1,200 for training and testing, respectively, following the original split~\cite{chen2024cariesxrays} for fair comparison.

\subsubsection{DVCT Dataset}
We collected a new benchmark dataset, named the DVCT dataset, which has a total number of 500,000 panoramic X-ray images, acquired from eight clinical centers, of which 498,000 remain unlabeled and 2,000 are annotated following the cross-verification by four experienced dental radiologists with over ten years of expertise, covering 5311 instances of dental caries and across dentition at different stage, as shown in Fig. \ref{fig:pipeline}(1). 
The labeled set is randomly divided into 1500/300/200 for training, validation and testing, respectively.
Our new dataset has two advantages over existing ones: 1) a higher-quality golden-standard annotated with dual verification from both panoramic X-ray images and intra-oral images and also a broader coverage of subjects across different age groups and 2) a large-scale unlabeled dataset, enabling self-supervised pretaining, especially for foundation model, which can be applied for any downstream task related to dental panoramic X-ray analysis.

\subsubsection{Evaluation Metrics}
For a consistent comparison, our evaluation primarily focuses on reporting the average precision (\%) across all benchmark datasets following previous work~\cite{chen2024cariesxrays}, where we adopt the standard AP metrics under various Intersection over Union (IoU) thresholds, ranging from 0.5 to 0.95.

\subsection{Implementation Details}
We implemented our dual-view architecture using ViT-B (14×14 patch size) for both branches. The global view pretraining utilized 518×518 global and 224×224 local crops, while the tooth-view operated at 112×112 global and 98×98 local crops to capture fine-grained dental structures. For caries detection, we employed ViTAdapter~\cite{chen2022vision} for multi-scale feature extraction at pretraining resolution, while maintaining a frozen pretrained ViT encoder with feature projection alignment in the tooth-view branch. 
The whole network was trained with AdamW optimizer where learning rate, weight decay, training epochs, and batch size were set to 0.0001, 0.05, 50, and 4, respectively. All experiments were conducted on NVIDIA A100 GPUs (80GB).

\subsection{Comparison with SOTA Approaches}

\begin{table}[t!]
\centering
\caption{Quantitative results on AAAI dataset and our DVCT dataset.}
\label{tab:detection_comparison}
\resizebox{\columnwidth}{!}{%
\begin{tabular}{l|c|ccc|ccc}
\hline
\multirow{2}{*}{\textbf{Methods}} & \multirow{2}{*}{\textbf{Backbone}} & \multicolumn{3}{c|}{\textbf{AAAI Dataset}}              & \multicolumn{3}{c}{\textbf{DVCT Dataset}}               \\ \cline{3-8} 
                                  &                                    & \textbf{AP}   & \textbf{AP$_{50}$} & \textbf{AP$_{75}$} & \textbf{AP}   & \textbf{AP$_{50}$} & \textbf{AP$_{75}$} \\ \hline
RetinaNet  ~\cite{lin2017focal}                       & ResNet-50                          & 13.0          & 30.5               & 10.2               & 11.1          & 32.9               & 3.3                \\
YOLOX    ~\cite{yolox2021}                         & CSPDarkNet                         & 40.5          & 81.3               & 36.1               & 15.4          & 42.4               & 8.4                \\
DINO     ~\cite{zhang2022dino}                         & Transformer                        & 37.8          & 75.3               & 29.4               & 22.2          & 50.4               & 14.4               \\
Faster RCNN   ~\cite{ren2015faster}                    & ResNet-50                          & 39.9          & 78.0               & 37.8               & 14.3          & 30.8               & 9.8                \\
FPCL      ~\cite{chen2024cariesxrays}                        & ResNet-50                          & 48.2          & 84.1               & 50.6               & 17.0          & 42.7               & 10.2               \\ \hline
\textbf{DVCTNet}                  & DINOv2-ViT-B                       & \textbf{48.9} & \textbf{84.7}      & \textbf{52.2}      & \textbf{31.3} & \textbf{57.4}      & \textbf{31.9}      \\ \hline
\end{tabular}
}

\end{table}
We compare our DVCTNet with the following typical object detection models:
RetinaNet~\cite{lin2017focal}, a single-stage detector that addresses class imbalance with focal loss;
YOLOX~\cite{yolox2021}, an anchor-free detector that achieves strong performance with a decoupled head;
DINO~\cite{zhang2022dino}, a DETR-like transformer-based detector with denoising training;
and Faster RCNN~\cite{ren2015faster}, a classical two-stage detector with a region proposal network.
We also compare with the recent SOTA detection method FPCL~\cite{chen2024cariesxrays}, specifically designed for dental caries detection.

\begin{figure}[t!]
    \centering
    \includegraphics[width=0.95\textwidth, height=140px]{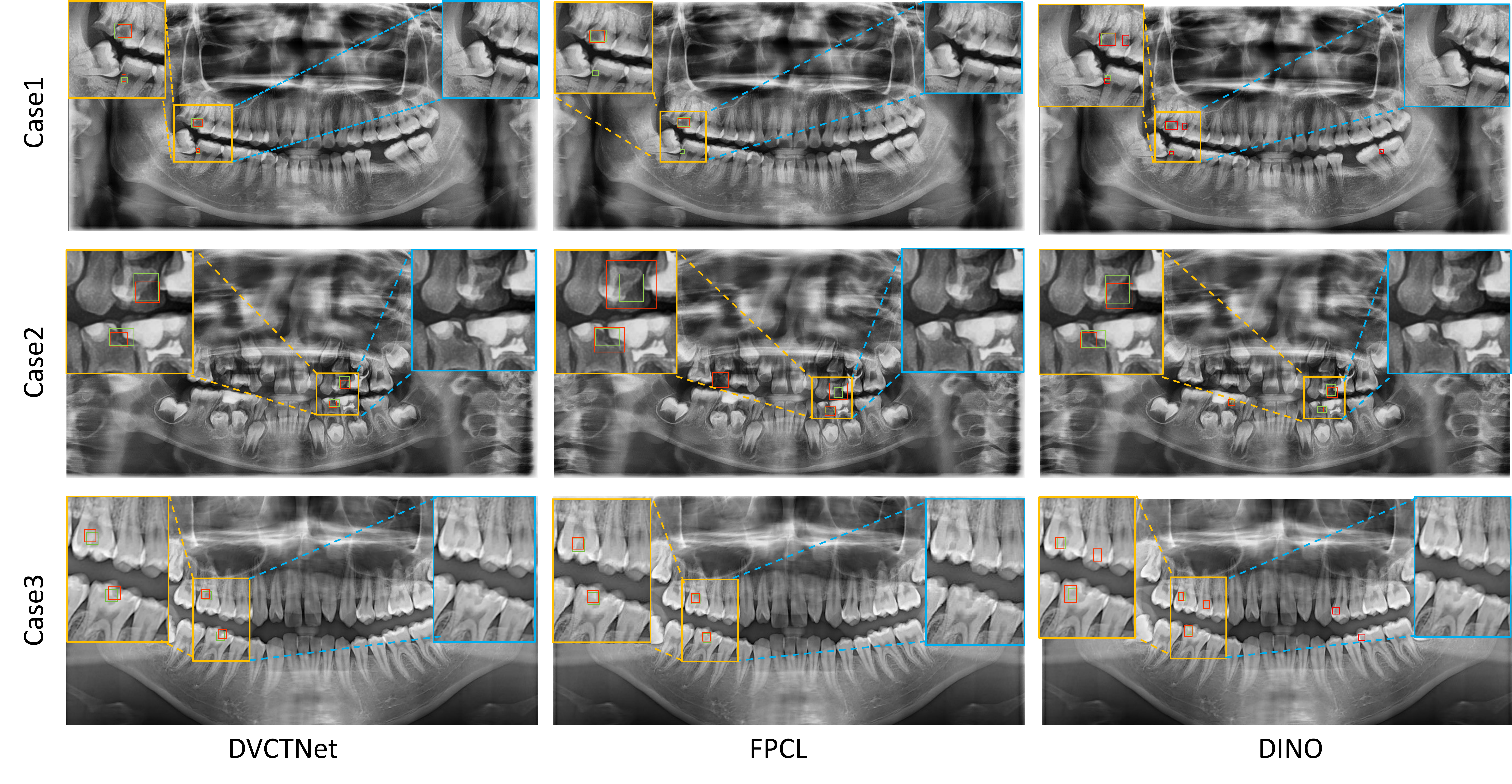}
    \caption{Visual comparison of dental caries detection methods. Dashed lines mark zoomed views: blue section shows the original image, orange section highlights the same area with overlaid ground-truth (green) and predicted (red) boxes.}
    \label{fig:result}
\end{figure}

Table \ref{tab:detection_comparison} presents the quantitative comparison results of different object detection methods on both the public dataset~\cite{chen2024cariesxrays} and our newly collected benchmark dataset. As evident, our DVCTNet consistently outperforms existing methods across all metrics. Specifically, on the AAAI dataset~\cite{chen2024cariesxrays}, DVCTNet achieves a slightly better result than the previous SOTA method FPCL by 0.7\%, 0.6\%, and 1.6\%, respectively. 
The performance gains are more significant on our dual-verified high-quality DVCT dataset, where DVCTNet attains 31.3\% AP, 57.4\% AP$_{50}$ and 31.9\% AP$_{75}$, demonstrating improvements of 14.3\%, 14.7\% and 21.7\% over FPCL and also suppresses the other detection methods at a large scale. 
Visualization results in Fig. \ref{fig:result} further demonstrate that our DVCTNet can detect tiny caries with low contrast variation across different scenarios. 
These substantial improvements further highlight the effectiveness of our core idea, which is to develop dual-view co-training that aligns with the clinical diagnosis workflow.

\subsection{Ablation Study}
\begin{table}[t!]
\centering
\caption{Ablation analysis for our proposed DVCTNet.} 
\label{tab:component_ablation}
\resizebox{\columnwidth}{!}{%
\begin{tabular}{cc|c|ccc|ccc}
\hline
\multicolumn{2}{c|}{\textbf{Dual-View Pretraining}} & \multirow{2}{*}{\textbf{GCV-Atten}} & \multicolumn{3}{c|}{\textbf{AAAI Dataset}} & \multicolumn{3}{c}{\textbf{DVCT Dataset}} \\ \cline{1-2} \cline{4-9} 
\multicolumn{1}{c|}{{~~~~~}\textbf{Global}{~~~~~}} & \textbf{Local} & & \textbf{AP} & \textbf{AP$_{50}$} & \textbf{AP$_{75}$} & \textbf{AP} & \textbf{AP$_{50}$} & \textbf{AP$_{75}$} \\ \hline
\multicolumn{1}{c|}{} & & & 32.8 & 64.3 & 28.5 & 15.7 & 33.0 & 12.1 \\
\multicolumn{1}{c|}{$\checkmark$} & & & 45.8 & 80.1 & 46.3 & 27.1 & 54.0 & 27.1 \\
\multicolumn{1}{c|}{$\checkmark$} & $\checkmark$ & & 46.2 & 81.7 & 47.8 & 27.8 & 54.6 & 28.1 \\
\multicolumn{1}{c|}{\textbf{$\checkmark$}} & $\checkmark$ & $\checkmark$ & \textbf{48.9} & \textbf{84.7} & \textbf{52.2} & \textbf{31.3} & \textbf{57.4} & \textbf{31.9} \\ \hline
\end{tabular}
}
\end{table}
We perform ablation experiments to analyze the effectiveness of each key component. As shown in Table~\ref{tab:component_ablation}, 
The baseline model without any of these components achieves an AP of 32.8\% and 15.7\% on the AAAI and DVCT datasets, respectively. Adding global-view pretraining significantly improves performance, with AP increasing by 13.0\% on AAAI and 11.4\% on DVCT. This significant gain demonstrates the importance of foundation model pretraining on panoramic X-rays for caries detection.
Incorporating the local-view pretaining model (which is implemented by simply concatenating $\mathbf{f}^{\mathrm{global}}$ and $\mathbf{f}^{\mathrm{local}}$ along the feature channel) enhances the results, with modest gains in AP on the two datasets, suggesting a better dual-view fusion strategy.
The addition of GCV-Atten yields the best performance across all metrics, achieving improvements of 2.7\% AP, 3.0\% AP$_{50}$, and 4.4\% AP$_{75}$ on the AAAI dataset, and similar gains on the DVCT dataset. These results validate that each component contributes positively to our framework.

\section{Conclusion}
In this work, we presented DVCTNet, a novel dual-view co-training framework for accurate dental caries detection that mimics the clinical workflow where dentists combine global panoramic screening with detailed tooth-level inspection. 
Our DVCTNet effectively integrates the complementary information from both views, leading to superior performance over existing SOTA methods.
We also established and released the first dual-verified benchmark dataset from intra-oral images and panoramic X-ray images for dental caries detection, named the DVCT dataset, which is the highest-quality dataset for this task so far, with the hope that it will provide a more comprehensive evaluation benchmark for the dental caries detection community.

%
\bibliographystyle{splncs04}
\bibliography{Paper-0146}
\end{document}